# Optical Generative Models

Shiqi Chen[a,b,c], Yuhang Li[a,b,c], Hanlong Chen[a,b,c], Aydogan Ozcan[a,b,c*]

[a]Electrical and Computer Engineering Department, University of California, Los Angeles, CA, 90095, USA

[b]Bioengineering Department, University of California, Los Angeles, CA, 90095, USA

[c]California NanoSystems Institute (CNSI), University of California, Los Angeles, CA, 90095, USA

Corresponding to: ozcan@ucla.edu

## Abstract

Generative models cover various application areas, including image, video and music synthesis, natural language processing, and molecular design, among many others. As digital generative models become larger, scalable inference in a fast and energy-efficient manner becomes a challenge. Here, we present optical generative models inspired by diffusion models, where a shallow and fast digital encoder first maps random noise into phase patterns that serve as optical generative seeds for a desired data distribution; a jointly-trained free-space-based reconfigurable decoder all-optically processes these generative seeds to create novel images (never seen before) following the target data distribution. Except for the illumination power and the random seed generation through a shallow encoder, these optical generative models do not consume computing power during the synthesis of novel images. We report the optical generation of monochrome and multi-color novel images of handwritten digits, fashion products, butterflies, and human faces, following the data distributions of MNIST, Fashion MNIST, Butterflies-100, and Celeb-A datasets, respectively, achieving an overall performance comparable to digital neural network-based generative models. To experimentally demonstrate optical generative models, we used visible light to generate, in a snapshot, novel images of handwritten digits and fashion products. These optical generative models might pave the way for energy-efficient, scalable and rapid inference tasks, further exploiting the potentials of optics and photonics for artificial intelligence-generated content.

# Introduction

Generative digital models have recently evolved to successfully create diverse, high-quality synthetic images [1-4], human-like natural language processing capabilities [5], new music pieces [6], and even new protein designs [7]. These emerging generative AI technologies are critical for various applications ranging from large language models (LLMs) [5] to embodied intelligence [8] and AI-generated content (AIGC) [9,10], which also gave rise to some of the well-known AI models, such as ChatGPT [11] and Sora [12]. Despite their success and generative power, these models are getting larger and larger, demanding significant amounts of electrical power, memory, and longer inference times [13]. The scalability and carbon footprint of such large generative AI models are also becoming a growing concern [14,15]. While several emerging approaches [16-38] have aimed to reduce the size and power consumption of such models, also improving their inference speed, there is still an urgent need to develop alternative approaches for designing and implementing power-efficient and scalable generative AI models.

Here, we demonstrate optical generative models that can optically synthesize monochrome or color images that follow a desired data distribution – *i.e.*, optically generating novel images that were never reported before for a given distribution. Inspired by diffusion models [4], this concept uses a shallow digital encoder to rapidly transform random two-dimensional (2D) Gaussian noise patterns into 2D phase structures that represent optical generative seeds. This optical seed generation is a one-time effort that involves a shallow and fast phase-space encoder acting on random 2D noise patterns. The optical generation of each novel image or output data following a desired distribution occurs on demand by randomly accessing one of these pre-calculated optical generative seeds. This broad concept can be implemented by different optical generative hardware, including, *e.g.*, integrated photonics- or free-space-based implementations as illustrated in **Fig. 1**. Without loss of generality, here we report free-space-based reconfigurable optical generative models (see **Figs. 1b, 2**). Each one of the optical generative seeds, once presented on a spatial light modulator (SLM) and illuminated by a plane wave, synthesizes a novel image through a reconfigurable diffractive decoder optimized for a given data distribution; the forward inference time of each novel image takes < 1 ns, and the refresh rate is limited by the frame-rate of the SLM that displays the pre-calculated optical generative seeds. This image generation process from the optical generative seed plane to the output image plane does not consume computing power except for the illumination wave. We report novel image generation performance that is statistically comparable to digital neural network-based generative models, which was confirmed through the generation of monochrome and multi-color novel images of handwritten digits, fashion products, butterflies and human faces, following the distributions of MNIST [46], Fashion MNIST [47], Butterflies-100 [51], and Celeb-A datasets [39].

To experimentally demonstrate the proof-of-concept of optical generative models, we built free-space-based hardware that works at 532 nm illumination in the visible part of the spectrum. Our experimental results demonstrated that the optical generative model can on-demand synthesize novel images in a snapshot by realizing the nonlinear domain mapping from 2D normal distribution to the target data distribution, covering novel images of handwritten digits (following the MNIST data distribution) and fashion products (following the Fashion MNIST data distribution). Our results confirmed that the learned optical generative models successfully grasped the underlying relationship within each target data distribution, creating novel images with an inference time of less than 1 ns for each optical generative seed that is randomly selected.

The presented framework is highly flexible since different generative optical models targeting different data distributions share the same optical architecture with a reconfigurable diffractive decoder. In this way, the desired data distribution can be switched from one generative AI task to another by simply changing the optical generative seeds that are pre-calculated from noise and the corresponding reconfigurable decoder surface, both of which can be performed without a change to the optical setup. We believe that the competitive performance, rapid inference speed, scalability and flexibility of optical generative models will stimulate further research and development, providing promising solutions for various applications of generative AI models, including, *e.g.*, AIGC, image and video processing and synthesis, among many others [40-45].

# Results

### Snapshot novel image generation using optical generative models

**Figure. 2** shows the schematic of our monochrome snapshot image generative model. As depicted in **Fig. 2a**, random 2D inputs, each following the normal distribution, are encoded into 2D phase patterns by a shallow digital encoder, which rapidly extracts the latent features and encodes them in the phase channel for subsequent wave propagation and analog processing. These random-noise-generated phase-encoded inputs serve as our optical generative seeds and are projected onto a spatial light modulator (SLM) to feed information to our free-space-based optical generative model. Under the illumination of coherent light, the optical fields carrying these encoded phase patterns propagate and are processed by a reconfigurable diffractive decoder that is optimized and fixed for a given target data distribution. Finally, the generated images are captured by an image sensor as intensity patterns, representing novel images following the target data distribution. The training procedure is shown in **Fig. 2b**, where we first train a proxy digital generative model based on Denoising Diffusion Probabilistic Model (DDPM) to learn the target data distribution [4]. Once trained, the learned DDPM is frozen, and it continuously generates the noise/image data pairs that are used to train the snapshot optical generative model. The shallow digital phase encoder and the optical generative model are jointly trained, enabling the model to efficiently learn the target distribution with a simple and reconfigurable architecture. **Figure. 2c** presents our blind inference procedure: the encoded phase patterns (*i.e.*, the optical seeds) generated by the digital encoder from random noise patterns are pre-calculated, and the optical generative model decodes these generative phase seeds through free-space, which significantly reduces inference time and energy consumption per novel image to be generated. For rapid synthesis of optical generative phase seeds from random Gaussian noise, the digital encoder comprised three fully-connected (FC) layers, where the first two are followed by a nonlinear activation function (see the Methods section for details). The reconfigurable diffractive decoder is optimized and structurally fixed for each target data distribution; stated differently, for different novel image generation tasks following different data distributions, it can be updated without changing the architecture or hardware of the optical generative model. This reconfigurable decoder is composed of a single surface with 800×800 learnable phase features, each covering 0-2$\pi$ range. More details about the snapshot image generation process are described in the Methods Section.

After separately training the corresponding model on the MNIST dataset [46] and the Fashion-MNIST dataset [47], we converged on two different optical generative models. In **Figs. 3a** and **b**, the snapshot generation of novel images of handwritten digits and fashion products, never seen before, are presented, showing high-quality output images for all data classes within these two data distributions. We used the Inception Score (IS) [48] and the Fréchet Inception Distance (FID) [49] as image quality metrics to evaluate the snapshot image generation performance of these optical generative models; see **Figs. 3c-d**. Both of these metrics are measured with a batch size of 1000 generated images, with the random integer seed $s \in [0, 10000)$ controlling the sampling of random Gaussian inputs $\mathcal{I}(x, y)$. In the IS evaluation, we generated the same number of novel images as the original dataset, aiming to measure the whole data distribution. To test the statistical significance of any potential improvement on the IS metric, we also carried out a t-test [56] between the optically generated image data and the original dataset. Based on this analysis, the higher IS values and the small p-values of <0.05 indicate that our snapshot optical image generative models produced statistically more diverse images compared to the original datasets. Using FID-based evaluations, we also show the statistics of 100 repeated calculations, demonstrating a high consistency between the optically generated images and the original data distribution.

To further evaluate the effectiveness of the snapshot optical generative models, we trained ten binary classifiers (based on a multi-layer perceptron architecture [50]) using the standard MNIST dataset, each tasked with recognizing a specific handwritten digit. These classifiers were used to assess 1,000 novel images (per class) created by our optical generative model. The assessment scheme and the results are presented in **Fig. 3e**. Overall, more than 94.2% of the handwritten digits generated by the snapshot optical generative model are correctly identified by the classifiers. Together with the outstanding IS and FID performance metrics of our optical generative model reported for each class in **Fig. 3g**, we realize that the snapshot optical generation produces images of new handwritten digits that follow the target distribution (revealed by the lower average FID) but have never appeared before in their style (as indicated by the higher average IS of the optically generated images). Hence, the digital classifiers trained on the original MNIST dataset exhibit some overfitting and face relative difficulty in distinguishing a small portion (< 5.8%) of the newly generated optical images of handwritten digits – which is expected.

Next, we evaluated the influence of the output diffraction efficiency ($\eta$) on the novel image generation performance of an optical generative model; $\eta$ is defined as the ratio of the total optical power distributed on the image sensor divided by the total input power illuminating the optical phase seed at the SLM plane. Depending on the illumination power available and the noise level encountered in the optical generative hardware, the diffraction efficiency of the system can be optimized by adding an $\eta$-related loss term during the training stage. By training several optical generative models targeting different levels of output diffraction efficiencies, we report in **Fig. 3f** an empirical relationship between FID and output diffraction efficiency of these models, measured with a batch size of 200 and repeated 100 times in different random seeds. Notably, for the optical generative models with a single decoding layer (blue line), the diffraction efficiency of the snapshot optical generative model can be increased to, *e.g.*, $\eta = 46.8\%$ on average with a minor compromise in the image generation quality, highlighting the capability of the snapshot optical generative model in achieving power-efficient image synthesis. We also trained additional optical generative models with five successive decoding layers (orange line), demonstrating a further improved image quality for a given level of output diffraction efficiency. These analyses indicate that, for a given image quality metric that is desired, a higher output diffraction efficiency can be achieved using a deeper decoder architecture, compared to the single-layer optical decoder. The phase structures of the jointly optimized decoding layers are also reported in **Supplementary Fig. S4b**.

We also conducted a performance comparison between the snapshot optical generative models and all-digital deep-learning-based models formed by the stacking of FC layers trained for the same image generation task. In **Fig. 4**, we show different configurations of these optical and all-digital generative models, where their computing operations (*i.e.*, the floating-point operations per second (FLOPs)), average IS values, and some examples of the generated novel images are reported for performance comparison of the two approaches. The comparative analysis in **Fig. 4** reveals that when the depth of the all-digital deep-learning-based image generative model is shallow, the output image quality cannot capture the whole distribution of the target dataset, resulting in failures or repetitive generations. On the other hand, the snapshot optical generative model with a shallow digital encoder is able to realize a statistically comparable novel image generation performance, matching the performance of a deeper digital generative model stacked with nine FC layers (see **Fig. 4**). This comparison further reveals that the snapshot optical generative model can reduce the needed number of FLOPs by approximately 3-fold with comparable generation of novel images matching the performance of deeper digital models.

We also compared the architecture of our snapshot optical generative models with respect to a free-space propagation-based optical decoding model, where the diffractive decoder was removed; see **Supplementary Figs. S1a-b**. The results of this comparison demonstrate that the reconfigurable diffractive decoder surface plays a vital role in improving the visual quality of the generated novel images. We also analyzed in **Supplementary Figs. S1a** and **c** the class embedding feature in the digital encoder; this additional analysis revealed that the snapshot image generation quality of an optical model without class embedding is lower, indicating that this additional information conditions the optical generative model to better capture the overall structure of the underlying target data distribution.

To further shed light on the physical properties of our snapshot optical generative model, in **Supplementary Fig. S2a** we explored the empirical relationship between the resolution of the optical generative seed phase patterns and the quality of the generated novel images. As the spatial resolution of the encoded phase seed patterns decreases, the quality of the novel image generation degrades, revealing the importance of the space-bandwidth product at the phase-encoded generative optical seed.

We also investigated the significance of our diffusion model-inspired training strategy for the success of snapshot optical generative models (see **Supplementary Fig. S2b**). When training an optical generative model as a generative adversarial network (GAN) [1] or a variational autoencoder (VAE) [2], we observed difficulty for the optical generative model to capture the underlying data distribution, resulting in a limited set of outputs that are repetitive or highly similar to each other – failing to generate diverse and high-quality novel images following the desired data distribution.

**Multi-color optical generative models**

We further extended the optical generative models for multi-color novel image generation using three different illumination wavelengths (*i.e.*, $\lambda_R$, $\lambda_G$, $\lambda_B$). **Figure. 5a** shows the schematic of our multi-color optical generative model, which shares the same hardware configuration as the monochrome counterpart reported earlier in the former section. In the multi-color novel image generation case, random Gaussian noise inputs of the three channels are also fed into a shallow and rapid digital encoder, and the resulting phase-encoded generative seed patterns at each wavelength channel are normalized ($\phi_{\lambda_R}$, $\phi_{\lambda_G}$, $\phi_{\lambda_B}$) to be sequentially projected ($t_1$, $t_2$, $t_3$) onto the same input SLM. Under the illumination of corresponding wavelengths in sequence, multi-color novel images following a desired data distribution are generated through a fixed diffractive decoder that is jointly optimized for the same image generation task. The resulting multi-color images are recorded on the same image sensor as before (see the Methods section for details).

We numerically tested this multi-color optical image generation framework using three different wavelengths ($473 nm$, $532 nm$, $633 nm$), where two different generative optical models were trained on the Butterflies-100 dataset [51] and the Celeb-A dataset [39] separately. Because these two image datasets do not have explicit categories, the shallow digital encoder only used randomly sampled Gaussian noise as its input without class label embedding. **Figures. 5b** and **c** show various novel images of butterflies and human faces produced by these multi-color optical generative models, revealing high-quality output images with various image features and characteristics that follow the corresponding data distributions. In **Figs. 5d** and **e**, the FID and IS performance metrics on Butterflies-100 and Celeb-A datasets are presented. The IS metrics and the t-test results show that the optical multi-color novel image generation model provides a statistically significant improvement ($p < 0.05$) in terms of image diversity and IS scores compared with the original Butterflies-100 dataset, whereas it does not show a statistically significant difference compared with the original Celeb-A data distribution. Additionally, some failed image generation cases are highlighted with red boxes in the bottom-right corner of **Figs. 5b** and **c**; such image generation failures were observed in 2.1% and 4.3% of the optically generated images for the Butterflies-100 and Celeb-A datasets, respectively. As illustrated in **Fig. 5f**, this image generation failure becomes more severe for the optical generative models that are trained longer. This behavior is conceptually similar to the mode collapse issue sometimes observed deeper in the training stage, making the outputs of the longer-trained multi-color optical generative models limited to some repetitive image features.

## Iterative optical generative models

The results and analyses performed so far were on snapshot optical generative models, where each phase-encoded optical generative seed was used to create a novel image through an optical decoder in a single snapshot illumination, with an image inference time of < 1 ns per wavelength channel. Following the iterative generative process of the original DDPM architecture, we also devised an iterative optical general model for *recursively* reconstructing the target data distributions from Gaussian noise [4]. As depicted in **Fig. 6a**, the iterative optical generative model also operates at three illumination wavelengths, where the multi-channel phase patterns encoded by the shallow digital encoder are sequentially projected onto the same SLM, as depicted earlier. To showcase the generative power of this iterative optical model, we used $L_o = 5$ decoding layers that are jointly optimized and fixed for the desired target data distribution. Different from the snapshot optical generative models discussed earlier, after recording the initial intensity image $\hat{\mathcal{I}}_t$ on the image sensor plane, the measured $\hat{\mathcal{I}}_t$ is perturbed by Gaussian noise with a designed variance, which is regarded as the iterative optical input $\mathcal{I}_{t-1}$ at the next timestep (timestep $t \in [0, T]$ and $\mathcal{I}_T \sim \mathcal{N}(0,1)$). The training process of such an iterative optical generative model is illustrated in **Fig. 6b**, where we sample a batch of timesteps $(t_1, t_2, \cdots)$ and accordingly add noise to the original data $\mathcal{I}_0$ to get the noised samples $(\mathcal{I}_{t_1}, \mathcal{I}_{t_2}, \cdots)$. These noised samples pass through the shallow digital encoder and the iterative optical generative model to get the successive outputs. Unlike the standard DDPM implementation, the iterative optical generative model directly predicts the denoised samples, with the loss function computed against $\mathcal{I}_0$ (see the Methods section for details). **Figure 6c** outlines the blind inference process of the iterative optical generative model, where the learned optical model recursively performs denoising on the perturbed samples from timestep $T$ to $0$, and the final generated novel image is captured on the sensor plane. Refer to the Methods Section for implementation details, and see **Supplementary Fig. S4d** for the resulting optical decoder structures.

Two different iterative optical generative models were trained for multi-color image generation following the distributions of Butterflies-100 [51] and Celeb-A [39] datasets. **Figures 7a** and **b** show the iterative optical generative model inference of novel images of butterflies and human faces. Compared to the multi-color generation of novel butterfly images using snapshot illumination per wavelength channel (reported in **Fig. 5b**), the iterative optical generative model produces higher-quality images. For the generation of novel images of human faces, following the data distribution of Celeb-A, the results of the iterative optical generative model also show more complex backgrounds, indicating an enhancement in the diversity of the novel image generation process. Another key advantage of the iterative optical generative models is that we do not encounter mode collapse throughout the training process. This is mainly because the successive iterations divide the distribution mapping task into independent Gaussian processes controlled by different timesteps, reducing the pressing demand on the snapshot image generation capability of the optical generative model.

To better highlight the vital role of the collaboration between the shallow digital encoder and the diffractive decoder in our iterative optical generative model architecture, we implemented an alternative iterative optical model trained on the Celeb-A dataset without using a digital encoder. As shown in **Fig. 7c**, this digital-encoder-free iterative optical generative model can also create multi-color images of human faces with different styles and backgrounds. This suggests that with the intensity-to-phase transformations directly implemented on an SLM without any encoder at the front-end along with the photo-electric conversion at the image sensor plane, we can achieve complex domain mappings with an iterative optical generative model – although with reduced performance and image diversity compared to the results of an iterative optical generative model that uses a digital encoder (**Fig. 7b**).

The intermediate results $\mathcal{I}_{t-1}$ ($t = 1000, 800, ..., 20, 1$) of the iterative optical generation models of **Fig. 7a-b** are also shown in **Fig. 7d**, which vividly demonstrates how the optical generative models gradually map the noise distribution into the target data domain. The FID and IS indicators of the iterative optical generative models are shown in **Figs. 7e** and **f**, respectively, where the details of these performance assessments were the same as those used in **Fig. 5d-e**. The results show an important improvement in the novel image generation performance of the iterative optical generative model, where the lower FID scores show that the generated novel images are closer to the target distribution. Furthermore, the higher IS values, along with the statistical t-test evaluations, indicate that the iterative optical generative model can create more diverse results than the original image dataset (*e.g.*, ~200,000 images in Celeb-A). We also report in the same figure the FID and IS values of the iterative optical generative model that was trained without a digital encoder, which shows a relatively worse performance compared to the iterative optical generative model jointly trained with a shallow digital encoder.

Next, we evaluated the impact of the iterative optical generative model's hyper-parameters on the output image quality through an ablation study; for this analysis, we selected an iterative optical generative architecture trained without a digital encoder, ensuring that only the optical parts contributed to the results. As shown in **Supplementary Figs. S2d-e**, the image generation quality decreases when the phase bit depth or the number of the decoder surfaces drops. We also investigated the influence of potential misalignments in the fabrication or assembly of a multi-layer diffractive decoder trained for iterative optical image generation. As shown in **Supplementary Fig. S3b**, lateral random misalignments cause a performance decrease in the novel image generation performance of an iterative optical model; however, training the iterative optical generative model with small amounts of random misalignments makes its bind inference more robust against such unknown, random perturbations, which is an important strategy to bring resilience for implementing deeper diffractive decoder architectures within an optical generative model.

## Experimental demonstration of snapshot optical generative models

We implemented an experimental demonstration of snapshot optical generative models using a reconfigurable system operating in the visible spectrum, as illustrated in **Fig. 8a** and **Supplementary Fig. S8**. The light from a laser (532 nm) is collimated to uniformly illuminate an SLM. The SLM displays the generative optical seeds containing the pre-calculated phase patterns $\phi(x, y)$ processed by a shallow and rapid digital encoder. After passing through a beam splitter, the optical fields modulated by the encoded phase patterns corresponding to the generative optical seeds are processed by a single-layer reconfigurable diffractive decoder. For each optical generative model, the state of the optimized decoder surface is fixed; and the same optical architecture is switched from one state to another, generating novel images that follow different target distributions. At the output of the snapshot optical generative model, the intensity of the generated novel images is captured by an image sensor (see **Fig. 8b** and the Method Section for details). A piston-based phase light modulator (PLM) is used as the reconfigurable decoding layer, with its diffractive features working like micro-mirrors, precisely moving back or forth along the optical axis.

For our experiments, we trained two different models for generating novel images of handwritten digits as well as fashion products, following the MNIST [46] and Fashion-MNIST datasets [47], respectively. **Figure. 8c** summarizes our experimental results for both models, demonstrating the successful generation of various novel handwritten digits and fashion product images. The successful generation of these novel images following the two target distributions highlights the robustness and versatility of the designed system, further validating the feasibility of snapshot optical generative models for a wide range of applications in image synthesis and generative AI. Since the encoded phase patterns can be pre-calculated by feeding random Gaussian noise to a shallow encoder network, they can be randomly accessed on demand, where novel images can be locally synthesized by the optical generative model, significantly saving decoding energy and time with an image inference time of <1 ns. Additional optically generated novel images of handwritten digits and fashion products are reported in **Supplementary Figs. S5-S6** and **Supplementary Videos 1-2**.

To further explore the latent space of the snapshot optical generative model, we also designed an additional optical experiment to investigate the relationship between the random noise inputs and the generated novel images. As shown in **Fig. 9a**, two random inputs $J^1$ and $J^2$ are sampled from the normal distribution $\mathcal{N}(0,1)$ and linearly interpolated using the equation $J^\gamma = \gamma J^1 + (1-\gamma)J^2$, where $\gamma$ is the interpolation coefficient. Note that the class embedding is also interpolated in the same way as the inputs. The interpolated input $J^\gamma$ and the class embedding are then fed into the trained digital encoder, yielding the corresponding generative phase seed, which is fed into the snapshot optical generative model hardware to output the corresponding novel image. **Figure 9b** presents the experimental results of this interpolation on the resulting images of handwritten digits using our optical generative setup. Each row displays novel images generated from $J^1$ (leftmost) to $J^2$ (rightmost), with intermediate images produced by the interpolated inputs as $\gamma$ varies from 0 to 1. The generated images exhibit smooth and coherent transitions between different handwritten digits, indicating that the snapshot optical generative model learned a continuous and well-organized latent space representation. Notably, the use of interpolated class embeddings demonstrates that the learned model realizes an external generalization: throughout the entire interpolation process, the generated novel images maintain recognizable digit-like features, gradually transforming one handwritten digit into another one through the interpolated class embeddings, suggesting effective capture of the underlying data distribution of handwritten digits. Additional interpolation-based experimental image generation results from our optical setup are shown in **Supplementary Fig. S7** and **Supplementary Videos 3-5**.

# Discussion

We demonstrated monochrome and multi-color optical generative models that share the same free-space-based configuration, composed of a shallow digital encoder, a phase SLM, and a reconfigurable diffractive decoder that remains fixed for each learned data distribution. The weights of the digital encoder and the diffractive decoder are jointly optimized to rapidly generate novel images following a target data distribution. Without changing this architecture or its physical hardware/setup, optical generation for a different data distribution can be implemented by simply reconfiguring the diffractive decoder to a new optimized state for decoding the generative phase seeds belonging to the new data distribution. This versatility of optical generative models might be especially important for edge computing, augmented reality (AR) or virtual reality (VR) displays, also covering different entertainment-related applications.

Using a snapshot optical generative model, we demonstrated novel image generation performance that is statistically comparable to deeper digital neural networks, also reducing the number of FLOPs by ~3-fold. Since the optical generative phase seeds can be pre-calculated from Gaussian noise and randomly accessed on demand, the local image inference delay and energy consumption from the encoded phase plane to the image plane are minimal.

Our results also showed that these optical generative models can be extended to high-quality image generation tasks. Guided by a proxy DDPM, the knowledge of the target distribution can be distilled to the multi-color optical generative model, as illustrated by the novel butterfly and human face images synthesized by our optical generative models. Furthermore, mimicking the diffusion process, our iterative optical generative models can learn the target distribution in a self-supervised way, which avoids mode collapse, generating even more diverse results than the original dataset (~200,000 images in Celeb-A), as illustrated in **Fig. 7**. In addition, the iterative optical generative models have the potential to eliminate the use of a digital encoder (see *e.g.*, **Fig. 7c**) and produce diverse outputs following different data distributions. Therefore, optical generative models can realize high-complexity distribution mappings by making use of different training strategies.

Although our results and analyses so far only considered the generation of novel images at a single 2D output field-of-view, spatially- and spectrally-multiplexed optical generative models can also be designed for parallel generation of a large number of novel images across different spatial and spectral channels. Moreover, benefiting from the inherent advantages of diffractive decoders in the rapid processing of visual information, optical generative models can also be designed to realize novel 3D image generation across a volume, which could open up new opportunities for AR/VR and entertainment-related applications, among many others.

Besides the free-space implementation of optical generative models reported in our Results section, the same inverse design strategy can also be implemented for integrated photonics-based on-chip processors, as outlined in **Fig. 1c**, to perform 1D calculations constituting an integrated photonics-based generative model.

Despite some of these important advantages and unique features discussed earlier, there are also some challenges associated with optical generative models in general.

Especially for free-space-based implementations, potential misalignments and physical imperfections within the optical hardware/setup present challenges, which might degrade the novel image generation performance. Another challenge is the limited phase bit depth of the optical devices or surfaces that physically represent the generative optical seeds and the decoder layer(s), which might constrain the upper-performance limit of novel image representation and generation. To mitigate some of these physical challenges, we can integrate these limitations directly into the training process of the optical generative model, which will accordingly shape both the digital encoder and the diffractive decoder, making the in silico optimized system align well with the physical limits and the capabilities of the hardware of the optical generative model.

We believe that the competitive performance, speed, efficiency and flexibility of the presented optical generative models will stimulate further research and development at the intersection of optics/photonics and the generative AI field, providing promising solutions for a variety of emerging applications.

## Supplementary Information includes:

- Methods Section
- Snapshot optical image generation process
- Iterative optical generative models
- Training strategy for optical generative models
- Implementation details of snapshot and multi-color optical generative models
- Experimental setup
- Formulation of DDPM
- Derivation of the distribution transformation coefficient
- Details of the U-Net architecture used in the proxy DDPM
- Supplementary Figures S1-S9
- Supplementary Videos 1-5

**Fig. 1: Illustration of optical generative models.**

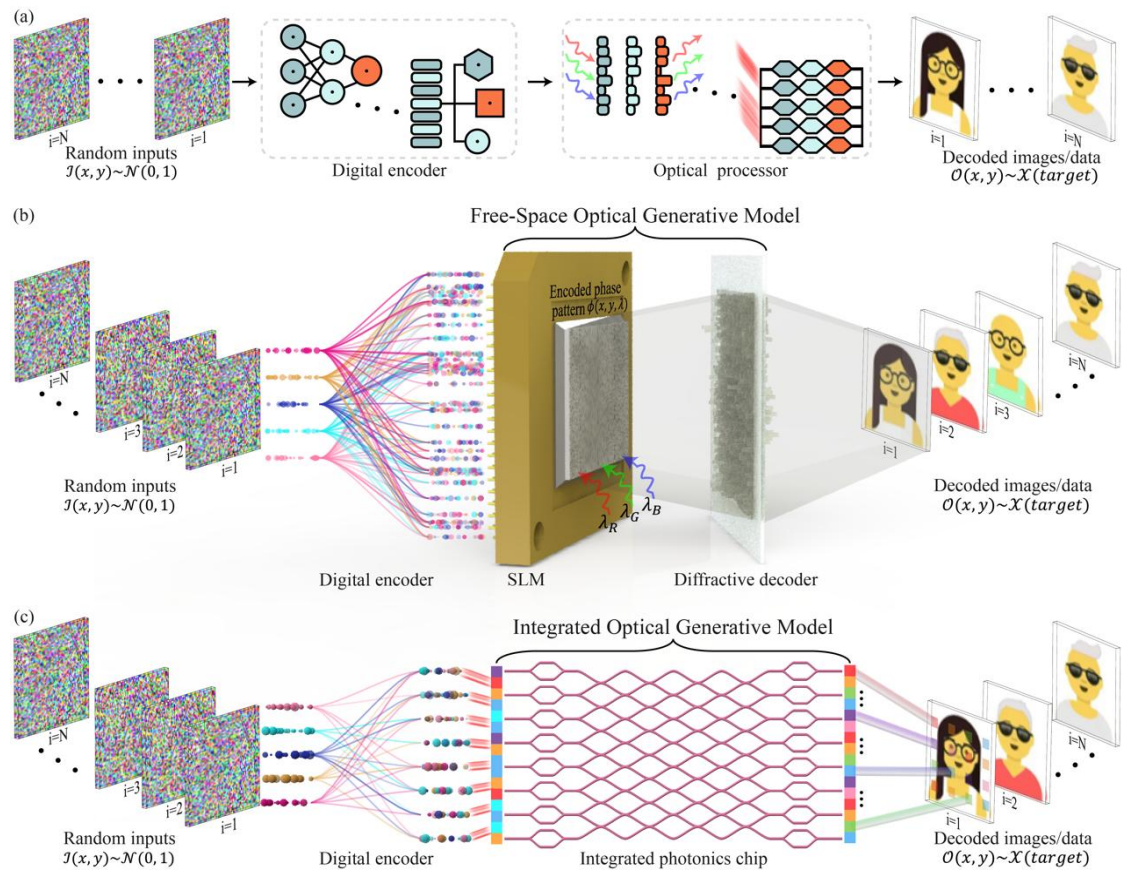

(a) General concept of optical generative models. Random inputs sampled from Gaussian noise are sequentially processed by a shallow digital encoder and an optical processor, generating novel output images (never seen before) following a desired data distribution – for example, generating novel images of human faces. (b) The design architecture of a free-space-based optical generative model. (c) Illustration of an integrated photonics chip-based optical generative model.

**Fig. 2: Design of a snapshot optical generative model**

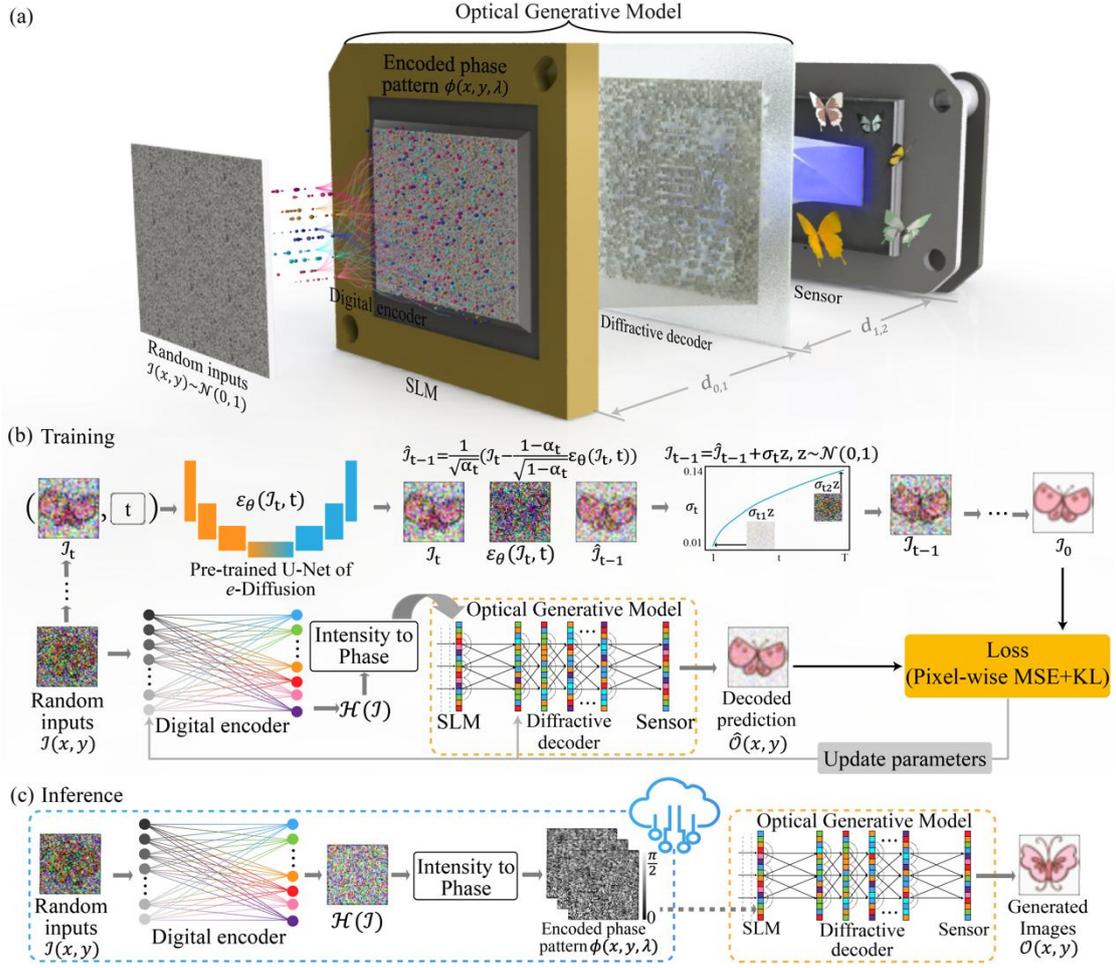

(a) Scheme of a snapshot optical generative model. Random Gaussian noise-based inputs are first encoded by a shallow digital encoder, which creates numerous optical generative seeds that are randomly accessed by an SLM. After the input optical field propagates through a reconfigurable and optimized diffractive decoder, the generated novel images are recorded on a sensor-array. For a given target data distribution, for example, images of butterflies or human faces, the generative optical model can synthesize countless images of novel butterflies or human faces, each with an inference delay of < 1 ns and without using computing power in the image synthesis process between the generative seed (SLM) plane and the image plane, except for the illumination light power. (b) The snapshot optical generative model is trained by a learned DDPM, where the data-pairs generated by the DDPM are used to guide the optimization of the snapshot optical generative model. (c) For blind inference of novel images, the pre-calculated optical generative seeds are randomly accessed through *e.g.*, a cloud-based server, where the snapshot image generation is locally realized by free-space optics and wave propagation.

**Fig. 3: Performance evaluations of snapshot optical generative models**

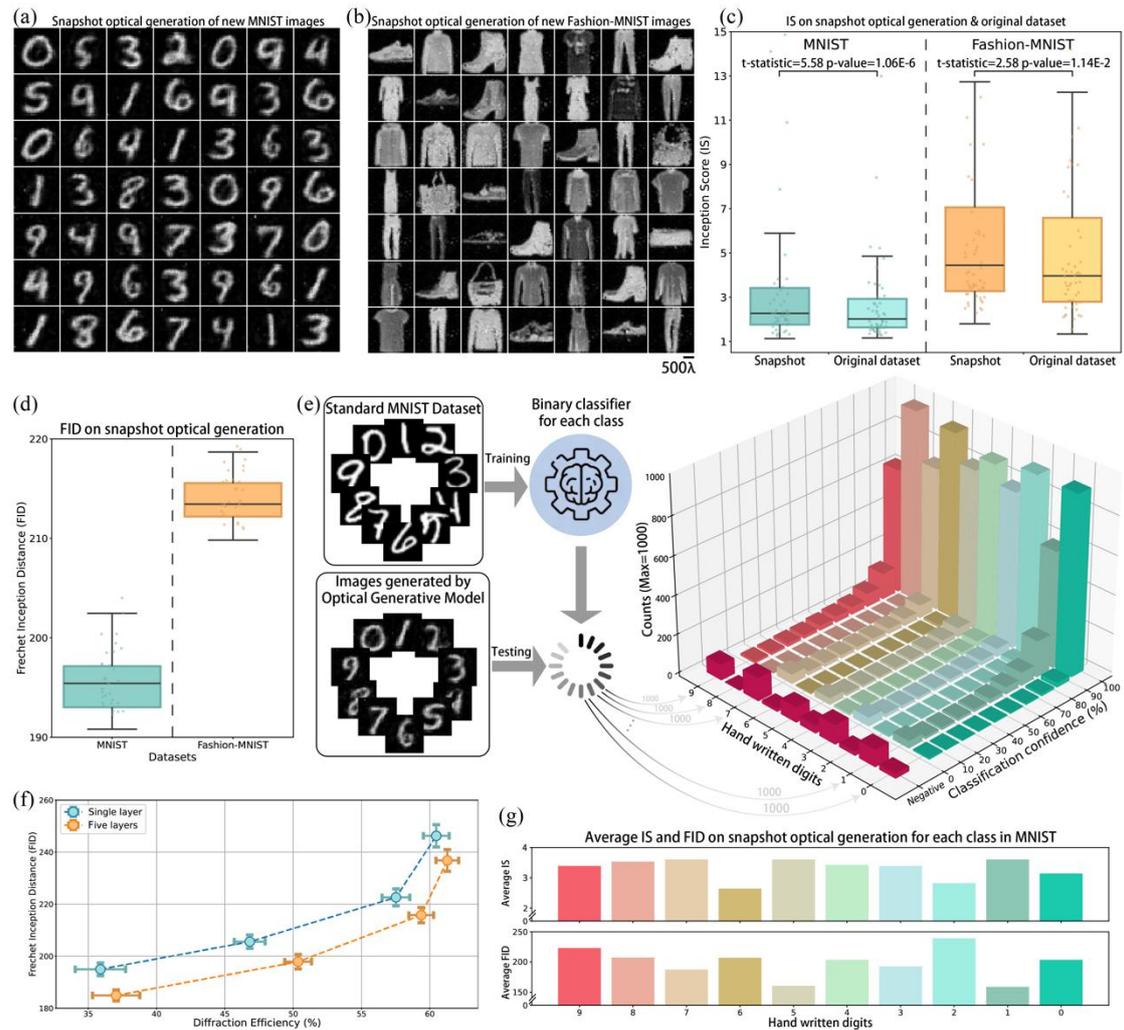

(a) Novel handwritten digit images (following the MNIST data distribution) generated by a snapshot optical generative model. (b) Novel fashion product images (following the Fashion-MNIST data distribution) generated by a snapshot optical generative model. (c) IS evaluation on snapshot image generation and the original target dataset, where the t-test results between two distributions are also reported. (d) FID assessment of MNIST and Fashion-MNIST snapshot optical image generation processes. (e) Classification confidence of 10 different binary classifiers (one for each digit), which were trained with the original MNIST data to test the quality of the novel images synthesized by the optical generative model. (f) Relationship between the FID scores and the output diffraction efficiency of optical generative models; one-layer vs. five-layer decoder-based snapshot optical image generation. (g) Average IS and FID metrics on snapshot optical image generation for each digit (9 to 0).

**Fig. 4: Performance comparison of snapshot optical generative models against digital-only generative models**

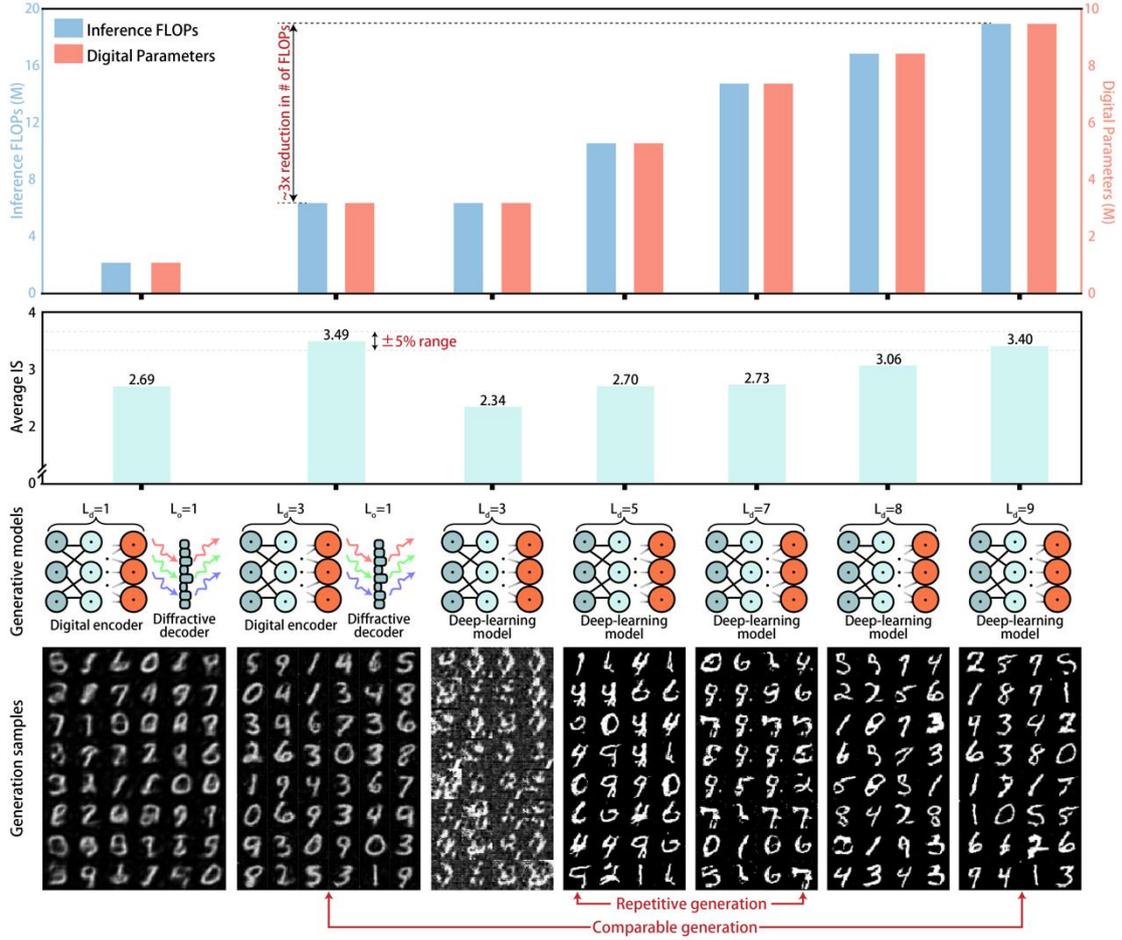

We report different configurations of generative models in the middle row. Left two models refer to snapshot optical generative models, where $L_d$ and $L_o$ are the number of FC digital layers and the number of diffractive decoder layers, respectively. Right five models refer to digital-only generative models with $L_d$ FC layers. In the bottom row, we show some examples of the generated images from these corresponding models. On the top 2 rows, we report the number of inference FLOPs, the number of digital parameters and the average IS values of the corresponding models. For a comparable novel image generation performance (average IS indicator), the number of digital FLOPs can be reduced by ~3-fold through the snapshot optical generative model. Note that repetitive image generations are observed for $L_d$= 5 and $L_d$= 7 digital-only models, which indicate restrictive image generation capability. Since such repetitively generated images might appear close to the target, the FID metric cannot truly reflect the generative capabilities of the model when it encounters such a mode collapse.

**Fig. 5: Multi-color optical generative models**

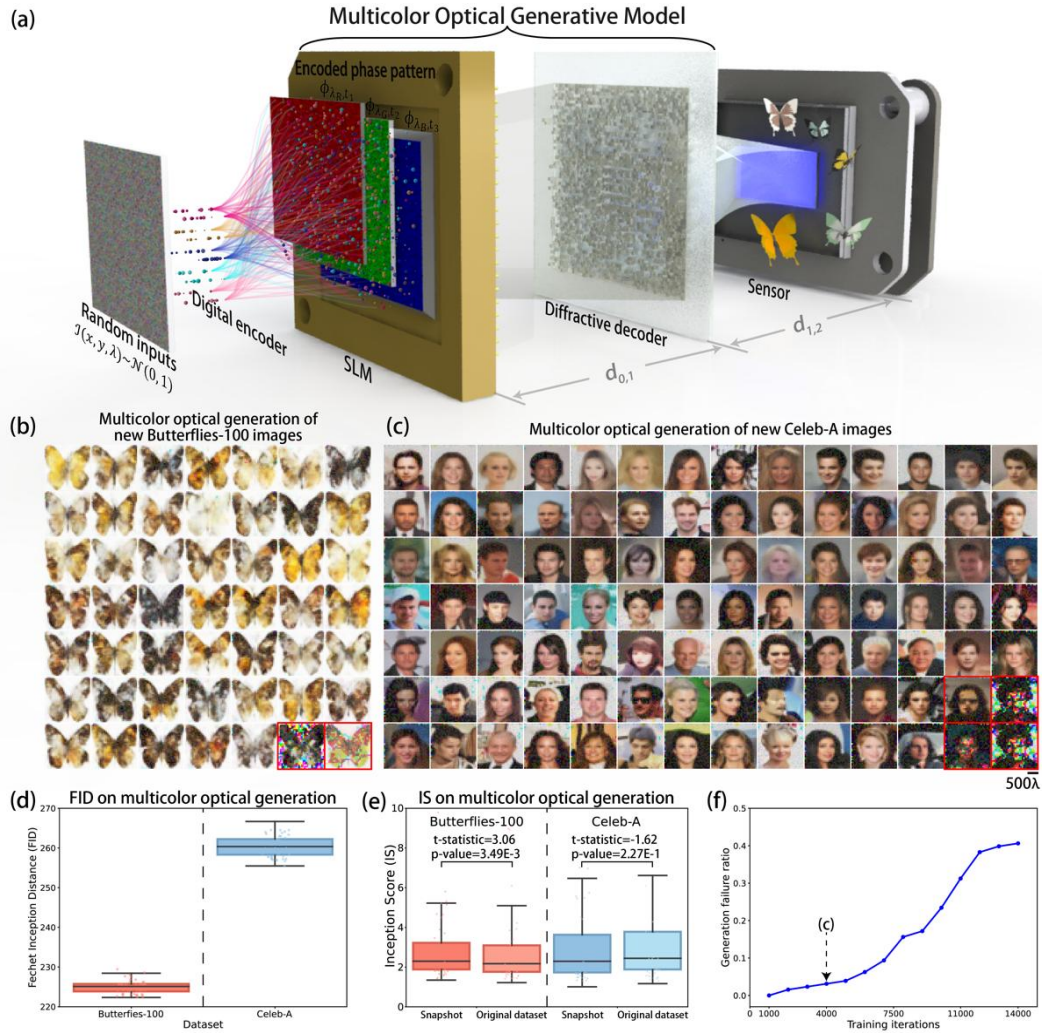

(a) Schematic of a multi-color optical generative model. (b) Novel butterfly images (following the Butterflies-100 data distribution) generated by a multi-color optical generative model. (c) Novel human face images (following the Celeb-A data distribution) generated by a multi-color optical generative model. We show some generation failure cases in the bottom-right of (b) and (c), with red frames. (d) FID evaluation of multi-color optical generative models. (e) IS evaluation of multi-color optical generative models against the original datasets, where the t-tests results between each pair of distributions are also listed. (f) The ratio of image generation failure as the training on Celeb-A continues. We marked the training iteration number of the multi-color optical generative model used in (c) with an arrow.

**Fig. 6: Iterative optical generative models**

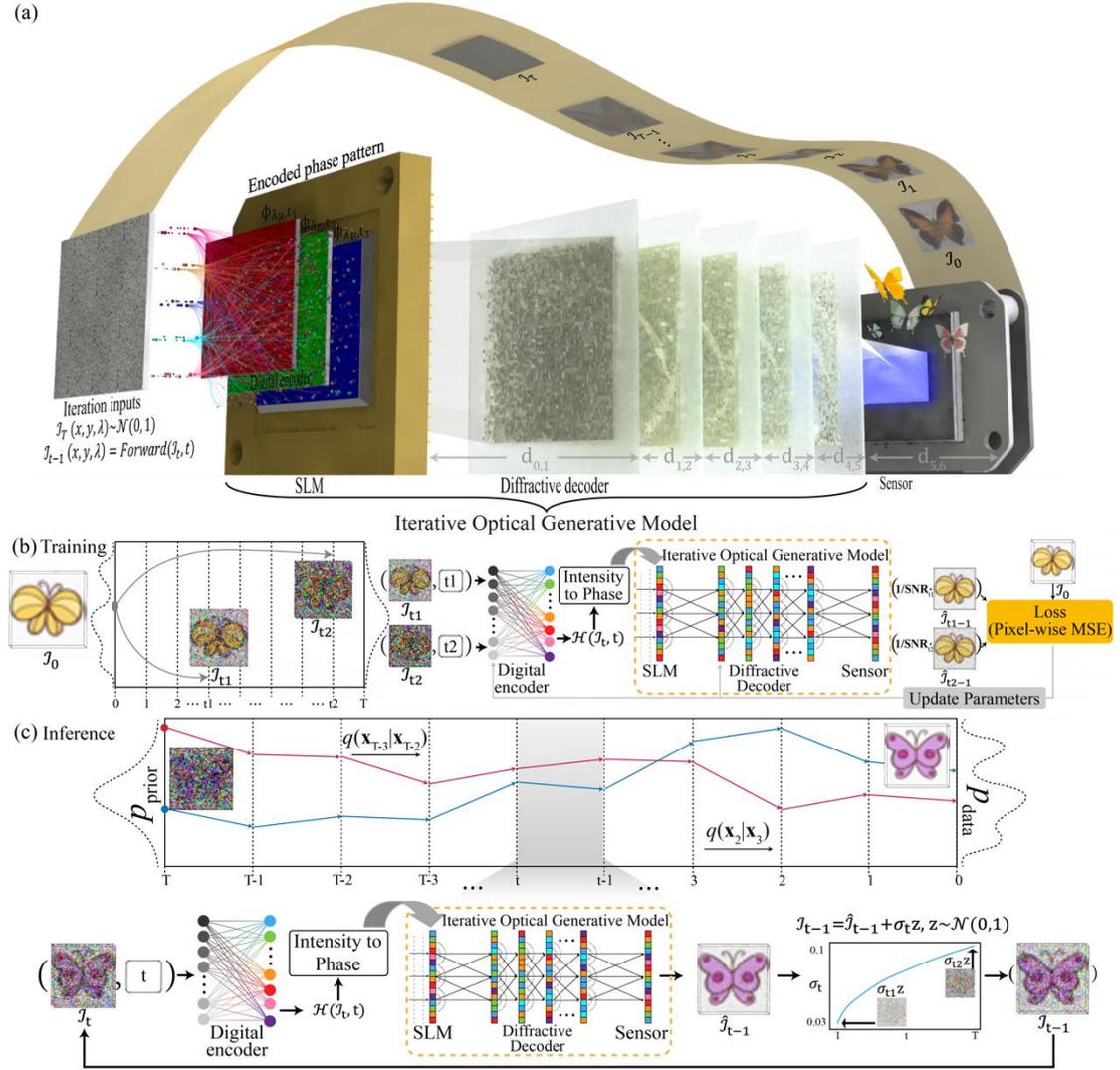

(a) Schematic of an iterative optical generative model. In each timestep, the noise-perturbed sample of the last timestep is input to the optical model. After the wave propagation, multi-color information is recorded to feed the next optical iteration with some scheduled noise added. For the last timestep, the image sensor-array records the output intensity for the final novel image generation. (b) The iterative optical generative model is trained using a digital proxy DDPM. (c) Following the training, in its blind inference, the iterative optical generative model gradually reconstructs the target data distribution (generating a novel image at timestep $0$) from Gaussian noise distribution (at timestep $T$).

**Fig. 7: Performance evaluations** of iterative optical generative models

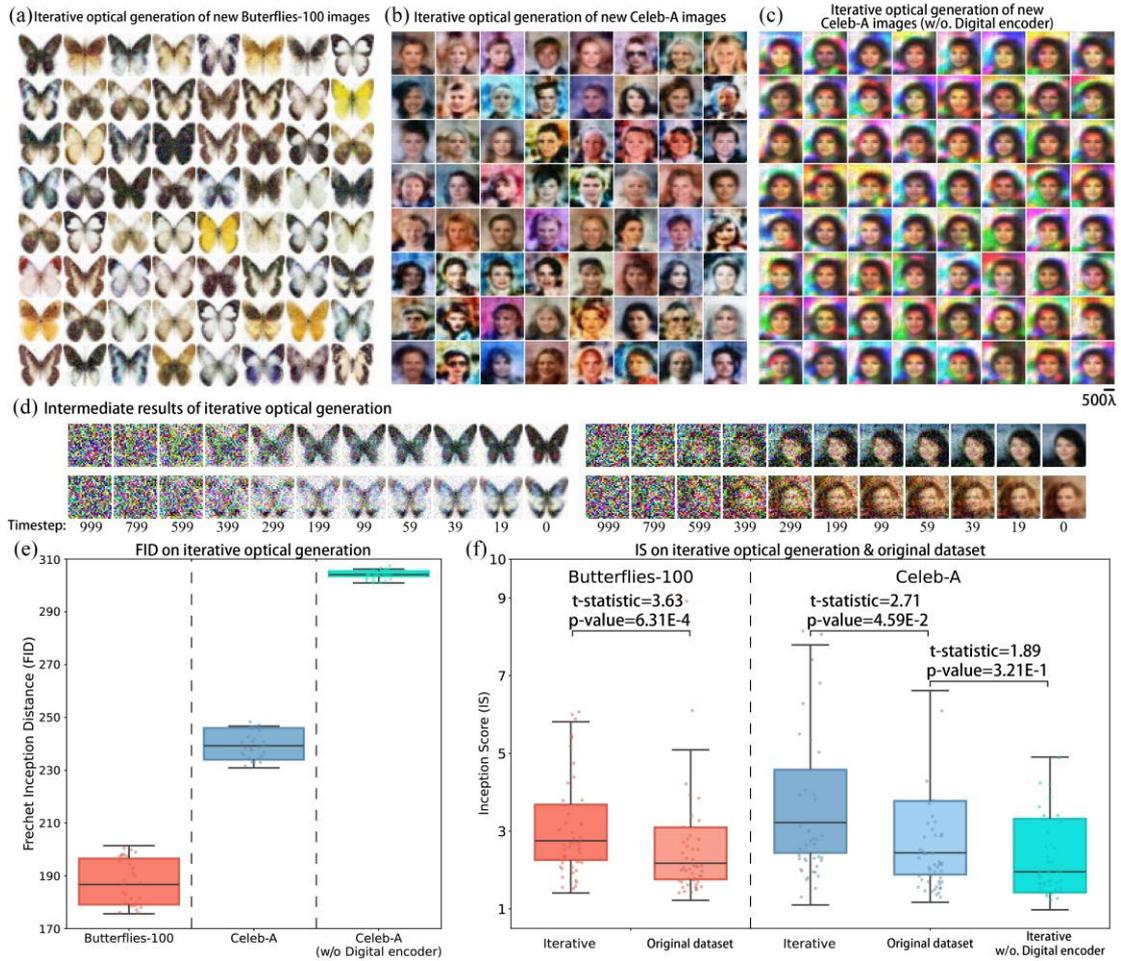

(a) Novel butterfly images (following the Butterflies-100 data distribution) generated by an iterative optical generative model. (b) Novel human face images (following the Celeb-A data distribution) generated by an iterative optical generative model. (c) Novel human face images generated by an iterative optical generative model that does *not* have a digital encoder. (d) Intermediate results of the iterative optical generative models at different timesteps. (e) FID assessment on iterative optical generation of novel butterfly and human face images. (f) IS comparisons of iterative optical generative model results and the original corresponding datasets (Butterflies-100 and Celeb-A). T-test results between each pair of distributions are also listed. The iterative optical generative models present higher IS values than the original datasets, which demonstrates that the optical models can generate more diverse images than the target data distributions.

**Fig. 8: Experimental demonstration of snapshot optical generative models**

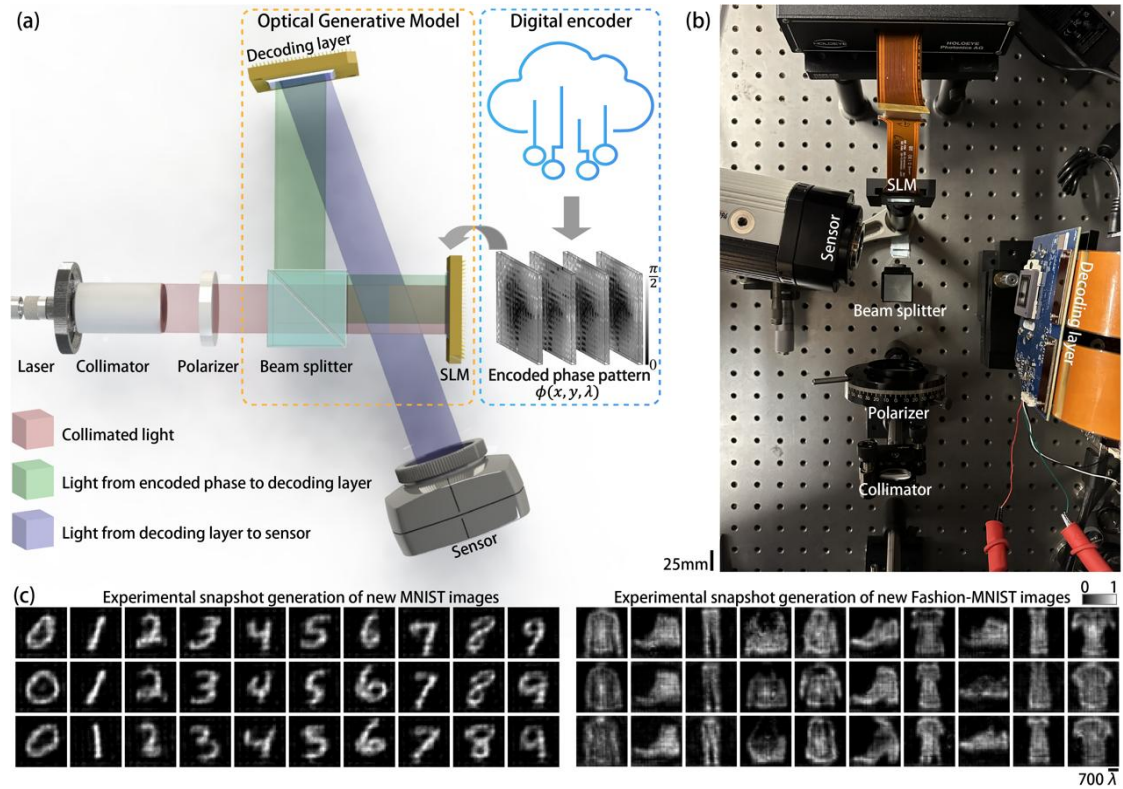

(a) Schematic of our experimental snapshot optical generative model. The encoded phase patterns, *i.e.*, the optical generative seeds, are pre-calculated and randomly accessed for each image inference task. (b) Photograph of the snapshot optical generative model. (c) The experimental results of novel image generation using the optical generative models trained for handwritten digits and fashion products, following the target data distributions of MNIST and Fashion-MNIST, respectively.

**Fig. 9: Experimental results of latent interpolation through a snapshot optical generative model**

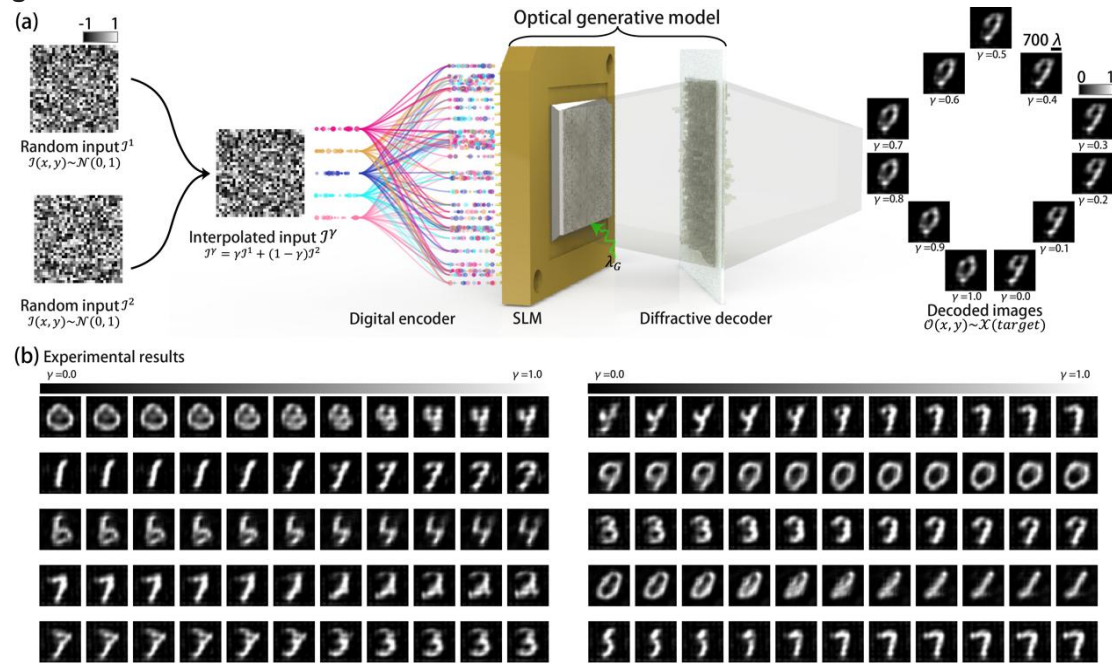

(a) We illustrate the latent interpolation carried out by a snapshot optical generative model, where two different random noise patterns (sampled from a normal distribution) and two class embeddings are first fused by weights, and then fed into the experimental optical generative model. The right part shows how the process of latent interpolation is controlled by the weights along with the interpolated class embeddings, gradually transforming one handwritten digit into another one. (b) More experimental results of latent interpolations are shown. Also see **Supplementary Fig. S7** and **Supplementary Videos 3-5**.